# Automatic segmentation of texts into units of meaning for reading assistance.

**Jean-Claude Houbart**
Pacte Novation
jchoubart@
pactenovation.fr

**Solen Quiniou**
LS2N
solen.quiniou@
univ-nantes.fr

**Marion Berthaut**
Mobidys
marion.berthaut@
mobidys.fr

**Béatrice Daille**
LS2N
beatrice.daille@
univ-nantes.fr

**Claire Salomé**
Pacte Novation
csalome@
pactenovation.fr

## Abstract

The emergence of the digital book is a major step forward in providing access to reading, and therefore often to the common culture and the labour market. By allowing the enrichment of texts with cognitive crutches, EPub 3 compatible accessibility formats such as FROG have proven their effectiveness in alleviating but also reducing dyslexic disorders. In this paper, we show how Artificial Intelligence and particularly Transfer Learning with Google BERT can automate the division into units of meaning, and thus facilitate the creation of enriched digital books at a moderate cost.

## 1. Introduction

Dyslexia is a specific learning disability that affects the acquisition of written language. We are studying developmental dyslexia that is not the result of a specific event such as a lesion. According to the French National Academy of Medicine, at least 3% to 5% of school-age children suffer from disorders (Expertise collective Inserm, 2007).

Although little studied, dyslexia also affects adults and seems to be a causal factor in illiteracy, which affects about 7% of the adult population who have been educated in France (Jonas, 2012)..  This creates difficulties employment for people affected with this condition.

Thus, in 1998, the Written and Oral Language Centre, attached to the scientific department of the IRSA (Interregional Institute for Health) studied a population of 124 young adults in difficulty of professional integration. 35% of the study population were very poorly or not at all able to read. Of these, half (17% of the studied population) had the characteristics of developmental dyslexia (Onisep, 2002).

In this context, the emergence of digital books represents a great hope for access to reading for all. The personalization and content enrichment capabilities of these tools open a new field to overcome decoding problems but also to treat them.

Digital books make it possible to enrich texts with various tools generally grouped under the term "cognitive crutches".  A first set of these crutches consists of features related to each word. The main ones are syllabic and phonemic coloration, highlighting of silent letters, audio support and tooltips.

In FROG format for example, a second set allows to go further in accompanying the reader and relies on the notion of rhesis. This concept is commonly used in speech therapy. It is empirically defined as "the amount of speech that can be delivered in an exhalation breath" (Brin, et al., 2016). The segmentation of a text into rhesis provides an intermediate scale between the word and the sentence that helps to structure the sentences for the application of several reading assistance tools. A good rhesis is generally a part of a sentence that is meaningful by itself.

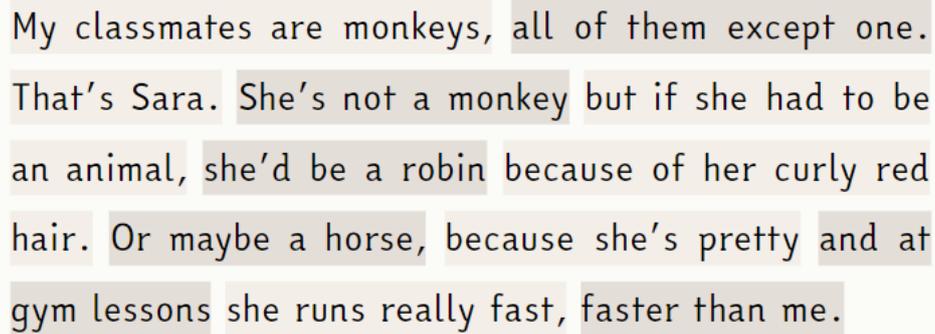

*Figure 1 : Example of Segmentation in rhesis*

The purpose of the research presented here is to use Artificial Intelligence techniques to automate the segmentation of texts, in order to reduce their cost and offer all audiences with reading difficulties the access to the largest number of books.

## 2. Utility of the rhesis

The origins of dyslexia are multiple and still subject to debate. Two theories dominate the study of dyslexic disorders: phonemic awareness (incorrect association of graphemes and phonemes) and visual processing (visual processing disorder prior to decoding).

Cognitive word crutches have proven their effectiveness (see (Snowling, 2000) for a review). The question here is to measure the contribution of a complementary segmentation into rhesis.

Segmentation in rhesis was studied in 2012 (Chilles Hélène, 2012). The comprehension of Literature or Mathematics texts was assessed for 9 students with dyslexia, 12 or 13 years old. The study concluded that the division into rhesis helps to lighten the working memory and seems to facilitate the processing of the information contained in the texts, including on mathematical statements.

(Labal & Le Ber, 2016) compared the contributions of an inverted prompter with a granularity per word or per rhesis. The study population consisted on 18 dyslexic children between 8 and 12 years of age. It appears that the word scale is on average the most relevant for reading performance (Error rate observed when reading aloud, reading speed). On the other hand, the standard deviation on the scores obtained on the rhesis segmentations was much greater than that on the word divisions: for some children, the rhesis division is more relevant. Above all, the granularity into rhesis was preferred by two thirds of the children. It seems that word division makes oral reading easier, but comprehension is helped by segmentation into rhesis. The two approaches are complementary and should be favoured according to each reader. The e-book responds well to this problem insofar as the granularity used can be left to the reader's choice.

In addition to the visual segmentation shown in Figure 1, segmentation in rhesis allows several cognitive crutches within a digital book:

- o The user can activate audio support: a unit of meaning is read when tapped.
- o A grey mask is applied to the text, a reading window highlights the unit of meaning that is pointed at.
- o Letters and words are spaced, the line spacing is increased, the paragraph is locked to the left and the units of meaning are not truncated.

## 3. Background

The first automatic segmentation tool was carried out in 2016 by the Natural Language Automatic Processing (NLP) team of the Laboratoire des Sciences du Numérique de Nantes (LS2N).

The methodology and results were presented in an article (Nin, et al., 2016). The segmentation principle is based on a sequence of sentence cutting operations until the span conditions (maximum length of the rhesis) are respected. The order of the segmentation criteria is as follows:

Punctuation => Proposals => Priority Prepositions => Syntagmas => Other Prepositions => Words.

Part-of-speech tagging and chunking are performed using OpenNLP, on models trained on the Free French Treebank (Hernandez & Boudin, 2013). The chunking tool is based on Markov model technology inspired by (Schmid & Atterer, 2004), supplemented by learning based on a text manually cut into rhesis. The computer language used is Java.

The first version of this tool, called RHEZOR, had a tendency to cut the text too finely. In 2018, a post-processing was added to group the rhesis together where possible while respecting the span.

Segmented texts and the e-book environment are provided by the French start-up Mobidys, which received the MGEN 2018 "Coup de Coeur" award for its contribution to the accessibility of school resources. Mobidys is supported by Pacte Novation, a company specialized in Artificial Intelligence, whose workforce is about 15% dedicated to research.

In 2019, with the help of the LS2N, the companies Mobidys and Pacte Novation decided to conduct a joint research project to develop a new version of the RHEZOR tool. The new prototype relies on the experience of the first version, while taking advantage of the sharp increase in the quantity of segmented texts available and the latest developments in terms of Transfer Learning in NLP.

## 4. Methodology

The graph shown in Figure 2 summarizes the three main milestones of our research work:

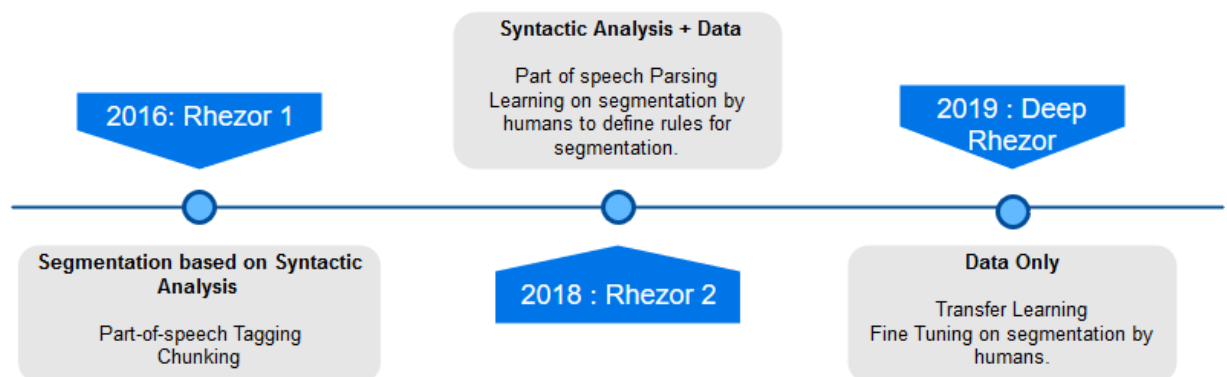

*Figure 2 : Main Steps of Development*

We implemented three versions of the Rhezor. Rhezor 1, presented in § 3, is mainly based on morpho-syntactic labelling and chunking processes. The small quantity of texts available at that time did not allow for learning.

In the second version initiated in 2018, the syntax analysis tool moves from OpenNLP to Spacy. On the other hand, a learning process was performed to define how to segment a sentence into rhetoric from the parsing. The operation is described in § 5.

Finally, we used the latest advances in transfer learning applied to natural language to learn Google's BERT model. This approach is called Deep Rhezor and is explained in § 6.

## 5. Rhezor 2: Syntactic analysis and learning

The SpaCy open source library (Honnibal, 2015) includes models that provide convolutional neural models for syntactic analysis and entity recognition. Spacy allows to analyze a text using a word prediction models. Each model is specific to a language and is driven by a data set. The model is pre-trained in 34 languages

In this way, SpaCy can identify, among other things, the grammatical nature of a word, or the links between the words in a sentence. A grammatical dependency tree can represent all this information (Figure 3. The nature of the word is in red. The dependency relationships between words is in blue).

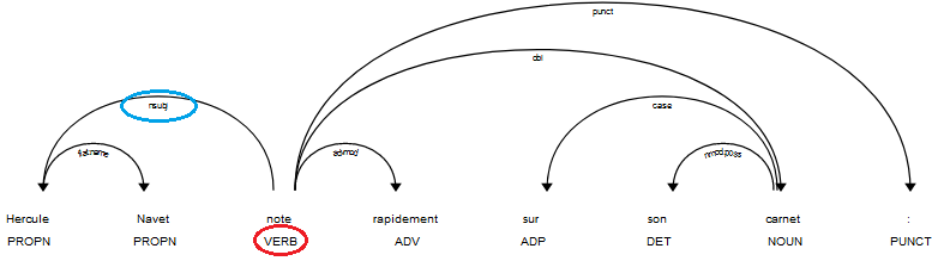

*Figure 3 . Dependency Tree generated by SpaCy for a sentence.*

Rhezor 2 uses the dependency tree of each sentence to perform a segmentation into rhesis. For this purpose, a score is calculated for each possible division according to the span. The score is defined according to the following criteria: type of segmented dependency; number of rhesis; balance in the lengths of the rhesis and deep of the segmentation in the tree.

An evolutionary algorithm determined the weighting between the criteria. It appears that the type of dependency is the main criterion. The Figure 4 compares the statistics on human and automatic segmentations.

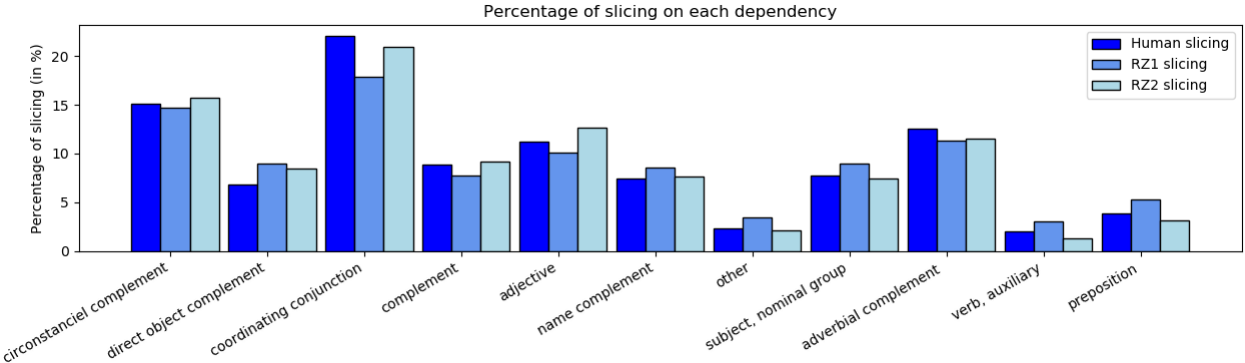

*Figure 4 : Statistical analysis of segmentations*

## 6. Deep Rhezor : Transfer Learning Only

Until recently, the volume of the corpus of manually segmented texts was far too small to consider a purely data-based approach. The progress made in 2018 on the attention mechanism and the concept of transformer (Vaswani, et al., 2017) now makes it possible to specialize a learned model with a reasonable volume of examples. This process, known as "Transfer Learning" has been used for several years for image recognition, but its application to natural language processing is very recent.

We selected the Google's BERT model (Devlin, et al., 2018), published in open source in October 2018. The learning data set is automatically generated from Wikipedia. First, about 15% of the words are hidden in each sentence to try to predict them. On the other hand, BERT has learned to predict whether two sentences are consecutive or not.

Google has generated several models. Two models were specifically generated for English and Chinese. Another multilingual model has also been created. Two models twice as large have also been tested for English and Chinese.

The texts base being mainly in French, we chose the model "BERT-Base, Multilingual Cased". The loss in accuracy on a translation task is about 3% compared to a model trained on a single language (at the same size). The maximum sentence size (max_seq_length) was fixed at 48 and the mini-lot size at 16. The Learning Rate was 2e-5 and the number of epochs was 3.

The implemented fine-tuning consisted in associating a sentence with one of its sub-sections, and creating a label indicating whether or not it is a rhesis. We had 10,051 sentences representing 53,478 rhesis, about one third of which was excluded of the training and kept for the assessment presented to the § 7. The general schema is proposed Figure 5.

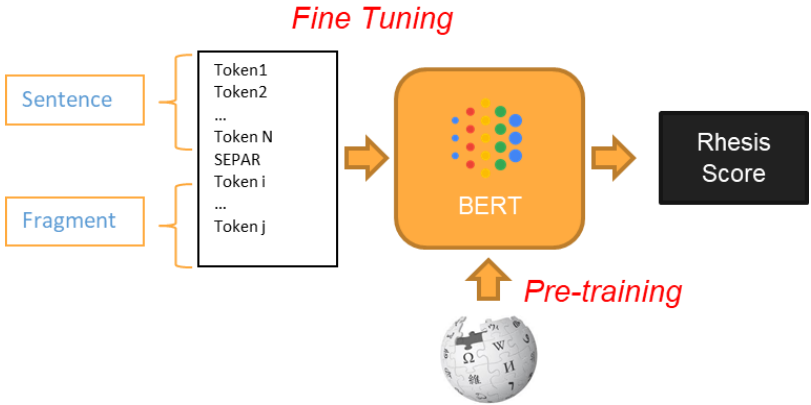

*Figure 5 : Fine Tuning Architecture*

## 7. Results

To assess the relevance of automatic segmentation, we used two means. The first, similar to precision, is to measure the percentage of the rhesis produced by the rhezor, which is also found in human segmentation. Note that this one is actually a correction of the output of Rhezor 1, which slightly penalizes the new versions. The second approach consists in segmenting a few pages taken at random from several books of different kinds, and having the result noted by people used to doing this work.

The Figure 6 summarizes the results of the first test. Rhezor 2 provides a 3.2% improvement over Rhezor 1. The Transfer Learning approach produces a 5% gain compared to Rhezor 2. We also observe that

Rhezor 1 remains first on Hercules Navet, probably because human segmentation was based on rhezor1. This advantage is not confirmed in the second evaluation.

| Text | Number of rhesis | Rhezor 1 Syntactic Analysis | Rhezor2 Syntactic Analysis and learning | Deep Rhezor Transfer learning |
|---|---|---|---|---|
| Le magicien Tre-Pi | 1633 | 58,0% | 65,7% | 68,7% |
| Hercule Navet | 761 | 76,7% | 71,3% | 72,7% |
| Emportée par le vent | 1805 | 65,7% | 71,2% | 72,2% |
| L'Avare | 7989 | 70,6% | 77,5% | 80,5% |
| Le Cid | 6670 | 77,6% | 75,6% | 85,5% |
| **Weighted Average** | **18858** | **71,8%** | **75,0%** | **80,1%** |

Figure 6: Percentage of common rhesis between human and automatic segmentation

In the second test, three Mobidys employees accustomed to segmentation of text into rhesis blindly evaluate the outputs of the three rhezors, as well as a human segmentation. Five pages of about 50 rhesis and randomly selected from the five evaluated books in the first test are used. Figure 7 shows the average score out of 20 for the different versions of the segmentation. We see that the ranking is identical to the one of the first test in Figure 6, with a steady progression between versions. In the case of the text extracted from the book "Le Cid", the average score for the division of the Deep Rhezor (Transfer Learning) exceeded the human segmentation (16/20 versus 15/20).

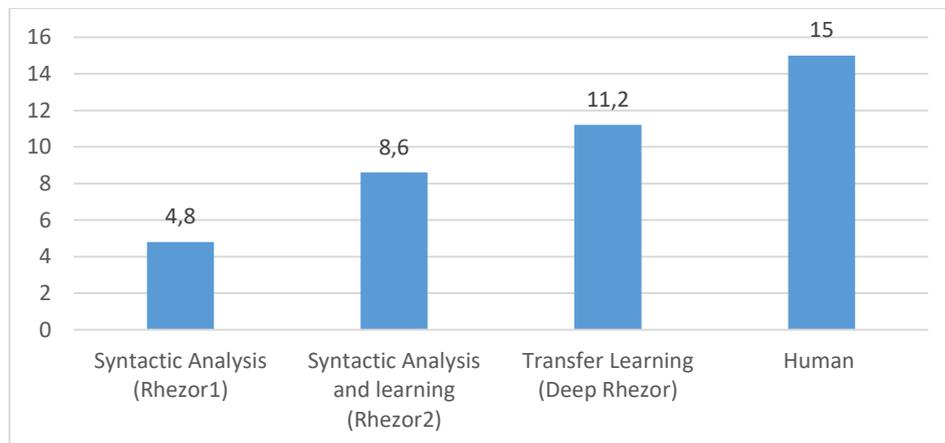

Figure 7: Human assessment of segmentations (average scores out of 20)

## 8. Conclusions

In this work, we have presented several approaches for automatic segmentation into rhesis. We can observe that Transfer Learning applied to NLP represents a significant step forward for our problem. In addition, there is room for improvement, by increasing calculation capabilities and the learning text base. Learning by text type (Youth, Novel, Theatre piece…) could also improve the quality of segmentation. We also plan then to work on the automatic generation of valid rhesis with a high level of probability and smart wrong rhesis in order to increase the volume and quality of data available for learning.

## 9. References

Brin, F., Courrier, C., LDerlé, E. & F., M., 2016. *Dictionnaire d'Orthophonie.* s.l.:Ortho Edition.

Chilles Hélène, M. B. S. F., 2012. *La mise en évidence des unités de sens d'un texte,* Académie de Strasbourg: Groupe recherche formation "Maîtrise de la langue et dyslexie, une gageure ?".

Devlin, J., Chang, M.-W., Lee, K. & Toutanova, K., 2018. *BERT: Pre-training of Deep Bidirectional Transformers for Language Understanding.* s.l.:Google AI Language.

Expertise collective Inserm, 2007. *Dyslexie Dysorthographie Dyscalculie Bilan des données scientifiques.* Paris: Les éditions Inserm.

Hernandez, N. & Boudin, F., 2013. *Construction automatique d'un large corpus libre annoté.* Paris, s.n.

Honnibal, M., 2015. *Introducing spaCy.* [Online] Available at: https://explosion.ai/blog/introducing-spacy

Jonas, N., 2012. Pour les générations les plus récentes, les difficultés des adultes diminuent à l'écrit, mais augmentent en calcul. *Insee Premiere*, Décembre.

Labal, P. & Le Ber, P., 2016. *Aide à la lecture des enfants dyslexiques dans le cadre de la création d'un livre numériques : effet de granularité par mot ou par unité de sens d'un prompteur inversé sur les performances et le confort de lecture.,* Université de Nantes: UFR Médecines et Techniques Médicales.

Nin, C., Pineau, V., Daille, B. & Quiniou, S., 2016. *Automatic segmentation of a text into rhesis.* Paris, s.n.

Onisep, 2002. Dyslexie et dysphasie. *Réadaptation,* Janvier.Issue 486.

Schmid, H. & Atterer, M., 2004. *New statistical methods for phrase break prediction.* Geneva, s.n.

Snowling, M. J., 2000. *Dyslexia.* Oxford, UK: Blackwell.

Vaswani, A. et al., 2017. *Attention Is All You Need.* s.l.:Google Brain, Google Research.